\documentclass[10pt,twocolumn,letterpaper]{article}

\usepackage[pagenumbers]{cvpr} 

\usepackage{wrapfig}

\newcommand{\bg}[1]{\boldsymbol{#1}} 
\newcommand{\bm}[1]{\mathbf{#1}} 

\newcommand\raiseT[2]{%
\setbox0\hbox{$#1{#2}$}\raise\dp0\box0}

\usepackage[dvipsnames]{xcolor}

\definecolor{cvprblue}{rgb}{0.21,0.49,0.74}
\usepackage[pagebackref,breaklinks,colorlinks,citecolor=cvprblue]{hyperref}

\def\paperID{288} 
\def\confName{3DV\xspace}
\def\confYear{2026\xspace}

\title{Adaptive Graph Kolmogorov-Arnold Network for 3D Human Pose Estimation}

\author{Abu Taib Mohammed Shahjahan and A. Ben Hamza\\
Concordia University, Montreal, Canada\\
{\tt\small abutaibmohammed.shahjahan@mail.concordia.ca}
}

\begin{document}
\maketitle

\begin{abstract}
Graph convolutional network (GCN)-based methods have shown strong performance in 3D human pose estimation by leveraging the natural graph structure of the human skeleton. However, their local receptive field limits their ability to capture long-range dependencies essential for handling occlusions and depth ambiguities. They also exhibit spectral bias, which prioritizes low-frequency components while struggling to model high-frequency details. In this paper, we introduce PoseKAN, an adaptive graph Kolmogorov-Arnold Network (KAN), framework that extends KANs to graph-based learning for 2D-to-3D pose lifting from a single image. Unlike GCNs that use fixed activation functions, KANs employ learnable functions on graph edges, allowing data-driven, adaptive feature transformations. This enhances the model's adaptability and expressiveness, making it more expressive in learning complex pose variations. Our model employs multi-hop feature aggregation, ensuring the body joints can leverage information from both local and distant neighbors, leading to improved spatial awareness. It also incorporates residual PoseKAN blocks for deeper feature refinement, and a global response normalization for improved feature selectivity and contrast. Extensive experiments on benchmark datasets demonstrate the competitive performance of our model against state-of-the-art methods. Code is available at: \textcolor{blue}{https://github.com/shahjahan0275/PoseKAN}
\end{abstract}

\section{Introduction}
3D human pose estimation aims to infer the 3D coordinates of body joints from 2D images or videos, a task that provides critical insights into human movement but remains challenging due to inherent depth ambiguities and occlusions in the input data. Mainstream approaches to 3D human pose estimation can be broadly categorized into one- and two-stage methods. The former aim to predict 3D joint coordinates directly from input images or video frames, bypassing any intermediate representation. However, they often struggle with occlusions, leading to inaccurate pose predictions~\cite{park20163d,sun2018integral,pavlakos2017coarse,sun2017compositional}. In contrast, two-stage methods decompose the problem into two separate stages: (1) detecting 2D joint locations in the image and (2) lifting these 2D detections into 3D space. The modular nature of two-stage pipelines helps optimize each component, selecting the best-performing 2D pose detectors and combining them with robust 2D-to-3D lifting models, ultimately leading to better 3D pose estimations and greater flexibility~\cite{martinez2017simple,ci2019optimizing,wu20203d,liu2020learning}. Recently, GCN-based methods have emerged as a powerful approach to 3D human pose estimation, effectively leveraging the inherent skeletal structure of the human body~\cite{zhao2019semantic,liu2020comprehensive,zou2021modulated,zou2021compositional,zhang2022group}. By formulating pose estimation as a graph-based learning problem, GCN-based models can effectively capture spatial dependencies between body parts. However, they still encounter several limitations that hinder their effectiveness in challenging 3D pose estimation scenarios. One of the primary limitations is their inability to model global context effectively due to their reliance on aggregating information mostly from immediate neighbors, thereby struggling to capture long-range dependencies between distant joints. They also incorporate Multi-Layer Perceptrons (MLPs) as their core components for feature transformation, inheriting a fundamental drawback of MLPs, namely spectral bias~\cite{Rahaman2019Bias}. As a result, these models tend to prioritize low-frequency components while struggling to capture high-frequency details, which are crucial for accurately modeling rapid movements and fine-grained joint interactions. This limitation can significantly impact pose estimation accuracy, particularly in highly articulated actions such as sports activities, where fine-grained motion details are essential for accurate predictions. Furthermore, GCN-based approaches lack interpretability due to their reliance on predefined activation functions. More recently, Bresson \textit{et al.}~\cite{Bresson2025KAGNNs} introduce KAGNNs, a family of graph neural networks that replace MLPs in standard message-passing layers with KANs for node feature updates, evaluating them on node and graph classification, link prediction, and graph regression.

\medskip\noindent\textbf{Proposed Work and Contributions.}\quad Addressing the aforementioned challenges of GCN-based methods for 3D human pose estimation requires a more adaptive and expressive approach that can dynamically adapt feature transformations, model long-range dependencies, and mitigate spectral bias. To this end, we propose an adaptive graph Kolmogorov-Arnold Network (PoseKAN), a novel graph-based framework that incorporates learnable function-based transformations. Drawing inspiration from the Kolmogorov-Arnold representation theorem~\cite{Braun2009KANTH,Johannes2021KANTH}, KANs~\cite{Liu2025KAN} have recently been introduced as a compelling alternative to MLPs, offering significant improvements in both interpretability and adaptability for function approximation tasks while mitigating spectral bias~\cite{Wang2025KANs}. The main contributions of our work can be summarized as follows:
\begin{itemize}
\item We present PoseKAN, a novel adaptive graph Kolmogorov-Arnold network framework for 3D human pose estimation. It employs \textit{2D-to-3D lifting from a single image} by levearing learnable function-based transformations, improving adaptability and generalization while maintaining computational efficiency.

\item We design a multi-hop feature propagation scheme that extends beyond one-hop aggregation by introducing a scaling parameter that controls the balance between local and global feature aggregation, improving robustness to occlusions and depth ambiguities.

\item We rigorously validate the effectiveness of our model through comprehensive experimental evaluation, including a comparative analysis and ablation studies against state-of-the-art methods on two benchmark datasets, demonstrating improved performance.
\end{itemize}

\section{Related Work}
Since our PoseKAN model follows the 2D-to-3D lifting pipeline~\cite{martinez2017simple,ci2019optimizing,liu2020learning,wu20203d}, we primarily focus in this section on relevant GCN-based methods within this category while also discussing how KANs fit into the broader landscape of graph-based learning.

\medskip\noindent\textbf{3D Human Pose Estimation.}\quad  The aim of 3D human pose estimation is to predict the 3D coordinates of body joints from 2D images or video frames. The 2D-to-3D lifting pipeline (i.e., two-stage) first detects 2D joint positions and then maps them to their 3D coordinates. Compared to one-stage methods, which directly regress 3D joint locations from images~\cite{sun2018integral,pavlakos2017coarse} without an intermediate representation, two-stage approaches have demonstrated greater robustness, particularly in handling occlusions and depth ambiguities~\cite{martinez2017simple,ci2019optimizing,liu2020learning,wu20203d}. Despite their success, most existing 2D-to-3D lifting methods utilize fully connected networks, which inherently lack spatial awareness and are prone to overfitting due to their densely connected architecture. Our proposed approach falls under the category of two-stage methods for \textit{2D-to-3D lifting from a single image}, focusing on learning flexible graph representations for more accurate 2D-to-3D pose lifting.

\medskip\noindent\textbf{Graph-based 3D Human Pose Estimation.}\quad Human body joints naturally form a structured graph, where nodes represent joints and edges capture relationships between them. Graph-based learning approaches, particularly GCNs~\cite{Kipf2017GCN}, have been widely applied to exploit this structured representation, showing promising results by leveraging the skeletal graph structure to model body joint dependencies~\cite{zhao2019semantic}. However, most GCN-based methods for 3D human pose estimation primarily aggregate information from one-hop neighbors, limiting their ability to model long-range dependencies. To address this, high-order GCN-based models~\cite{zou2020high,Shah2024FlexGCN} extended GCNs to multi-hop neighborhoods, capturing long-range dependencies beyond immediate joint connections. Modulated GCN~\cite{zou2021modulated} introduced adjacency matrix modulation, enabling the model to learn additional edges beyond the predefined skeletal structure. GroupGCN~\cite{zhang2022group} and Multi-hop Modulated GCN~\cite{lee2022multi} further refined graph feature propagation by fusing multi-hop neighborhood information. More recently, GraphMLP~\cite{li2025graphmlp} combined MLPs and GCNs to leverage both global and local skeletal interactions. While GCN-based approaches have led to notable improvements in 3D human pose estimation, they still face three key limitations. First, standard GCN architectures rely on one-hop neighborhood aggregation. This restricts the ability to capture long-range dependencies, which are crucial for handling occlusions and depth ambiguities. Second, like MLPs, GCNs exhibit spectral bias, meaning they favor low-frequency components while struggling to capture high-frequency details, thereby hindering their ability to accurately represent complex poses and rapid movement changes. Third, GCN-based methods apply predefined activation functions at the node level, restricting adaptability. By comparison, our proposed PoseKAN framework learns activation functions dynamically on graph edges, providing greater flexibility in feature transformation and improving interpretability, while mitigating the spectral bias thanks, in large part, to learning function-based transformations that help capture both low- and high-frequency components. Low-frequency components represent smooth, gradual variations in joint positions over time or space, whereas high-frequency components correspond to rapid, abrupt changes in motion. In the Human3.6M dataset~\cite{ionescu2013human3}, for instance, actions such as Walking and Eating predominantly exhibit low-frequency characteristics due to their smooth and predictable patterns. In contrast, high-frequency components are more prominent in actions like Greeting, where sudden movements and finer pose adjustments are required.

\section{Method}
In this section, we first describe the 3D human pose estimation task, followed by a preliminary background on KANs~\cite{Liu2025KAN,Wang2025KANs}. Subsequently, we introduce a novel spectral modulation filter, which serves as the cornerstone for our proposed adaptive graph KAN model.

\medskip\noindent\textbf{Problem Description.} \quad Let $\mathcal{G}=(\mathcal{V},\mathcal{E},\bm{X})$ be an attributed graph representing the skeletal structure of the human body, where $\mathcal{V}=\{1,\ldots,J\}$ is the set of $J$ nodes (i.e., body joints) and $\mathcal{E}\subseteq \mathcal{V}\times\mathcal{V}$ is the set of edges, and $\bm{X}$ a $J\times F$ feature matrix of node attributes. We denote by $\bm{A}$ a $J\times J$ adjacency matrix whose $(i,j)$-th entry is equal to 1 if $i$ and $j$ are neighboring nodes, and 0 otherwise. We also denote by $\hat{\bm{A}}=\bm{D}^{-\frac{1}{2}}\bm{A}\bm{D}^{-\frac{1}{2}}$ the normalized adjacency matrix, where $\bm{D}=\mathsf{diag}(\bm{A}\bm{1})$ is the diagonal degree matrix and $\bm{1}$ is a vector of all ones. Given a training set $\mathcal{D}=\left\{\left(\mathbf{x}_{i}, \mathbf{y}_{i}\right)\right\}_{i=1}^{N}$ comprised of 2D joint positions $\bm{x}_{i}\in\mathcal{X}\subset\mathbb{R}^2$ and their associated ground-truth 3D joint positions $\bm{y}_{i}\in\mathcal{Y}\subset\mathbb{R}^3$, the aim of 3D human pose estimation is to learn the parameters $\bm{w}$ of a regression model $f_\bm{w}: \mathcal{X} \rightarrow \mathcal{Y}$ by finding a minimizer of a loss function.

\medskip\noindent\textbf{Kolmogorov-Arnold Networks.}\quad KANs are inspired by the Kolmogorov-Arnold representation theorem~\cite{Braun2009KANTH,Johannes2021KANTH}, which states that any continuous multivariate function on a bounded domain can be represented as a finite composition of continuous univariate functions of the input variables and the binary operation of addition. A KAN layer is a fundamental building block of KANs~\cite{Liu2025KAN}, and is given by a matrix $\bg{\Phi}=(\phi_{q,p})$ of 1D functions, where each trainable activation function $\phi$ is defined as a weighted combination, with learnable weights, of a sigmoid linear unit (SiLU) function and a spline function:
\begin{equation}
\phi(x)=w_{b}b(x) + w_{s}\text{spline}(x),
\end{equation}
where $b(x)=\text{SiLU}(x)=x/(1+e^{-x})$ and $\text{spline}(x)=\sum_{i}c_{i}B_{i}(x)$ is a weighted sum of B-splines basis functions with trainable coefficients $c_{i}$. During training, the weights $w_{b}$ and $w_{s}$ are learned to optimize performance. Given a input feature vector $\bm{x}^{(\ell)}\in\mathbb{R}^{F_{\ell}}$, the output of the $\ell$-th KAN layer is an $F_{\ell+1}$-dimensional feature vector given by
\begin{equation}
\bm{x}^{(\ell+1)} = \text{KAN}^{(\ell)}(\bm{x}^{(\ell)}).
\end{equation}

\subsection{Spectral Modulation Filtering}
Spectral graph filtering employs filters defined as functions of the graph normalized Laplacian (or its eigenvalues). The goal of these filters, often referred to as frequency responses or transfer functions, is to reduce or eliminate high-frequency noise in the graph signal. These functions basically describe how a filter affects the input graph signal to produce the output graph signal. We define a spectral modulation filter as follows:
\begin{equation}
h_{s}(\lambda)=\frac{1}{(1+s)\lambda - s\lambda^2},
\end{equation}
where $s\in (0,1)$ is a positive scaling parameter that allows for the modulation or adjustment of the spectral characteristics, indicating its capability to control the filtering effect on different frequency components of the graph signal. The filter $h_s$ is a rational polynomial function of the eigenvalues of the normalized Laplacian matrix. It is an adaptive-pass filter in the sense that it exhibits low-pass characteristics, as it allows low-frequency components (corresponding to small eigenvalues) to adaptively pass through with little attenuation, while attenuating high-frequency components (associated with large eigenvalues). The attenuation behavior of the filter is determined by the scalar $s$, which serves as a tuning or modulation parameter. By adjusting $s$, we can control the trade-off between preserving low-frequency structural information and reducing high-frequency noise or variations in the graph signal. When $s$ is large, the filter is less selective, allowing a wider range of frequencies to pass through with less attenuation. As $s$ decreases, the filter becomes more selective and significantly attenuates higher frequencies, effectively filtering out more of the high-frequency noise or variations in the graph signal.

\medskip\noindent\textbf{Graph Filtering System.}\quad Applying the spectral modulation filter on the graph signal (i.e., feature matrix) $\bm{X}\in\mathbb{R}^{N\times F}$ yields a filtered graph signal $\bm{H}\in\mathbb{R}^{N\times F}$ given by
\begin{equation}
\bm{H}=h_{s}(\bm{L})\bm{X}=((1+s)\bm{L}-s\bm{L}^{2})^{-1}\bm{X},
\end{equation}
which can be rewritten as
\begin{equation}
((1+s)\bm{L}-s\bm{L}^{2})\bm{H}=\bm{X},
\end{equation}
or equivalently
\begin{equation}
\begin{split}
\bm{H} &=(\bm{I}-(1+s)\bm{L}+s\bm{L}^{2})\bm{H}+\bm{X} \\
       &=(\bm{I}-s\bm{L})(\bm{I}-\bm{L})\bm{H}+\bm{X}.
\end{split}
\end{equation}
Since $\bm{L}=\bm{I}-\hat{\bm{A}}$, where $\hat{\bm{A}}$ is the normalized adjacency matrix, the spectral modulation filter equation becomes
\begin{equation}
\bm{H}=((1-s)\bm{I}+s\hat{\bm{A}})\hat{\bm{A}}\bm{H}+\bm{X},
\label{Eq:SHA}
\end{equation}
which can be solved using, for instance, the fixed point iteration method~\cite{Burden2015Numerical}, where $\bm{H}=\varphi(\bm{H})$ with the function $\varphi$  defined by the right-hand side term of~\cref{Eq:SHA}.

\medskip\noindent\textbf{Fixed Point Iterative Solution.}\quad Fixed-point iteration is an iterative numerical method used to find a fixed point of a given function~\cite{Burden2015Numerical}. The process involves repeatedly applying the function to an initial guess or estimate and updating this estimate in each iteration until it converges to the fixed point. For the spectral modulation filter equation $\bm{H}=\varphi(\bm{H})$, the fixed point iterative solution is given by
\begin{equation}
\bm{H}^{(t+1)}=((1-s)\bm{I}+s\hat{\bm{A}})\hat{\bm{A}}\bm{H}^{(t)}+\bm{X},
\label{Eq:FP}
\end{equation}
with some initial guess $\bm{H}^{(0)}$, and $t\in\mathbb{N}$ denotes the iteration number.

\subsection{Proposed Graph KAN Model}
While GCN-based methods have shown strong performance in 3D human pose estimation, they, however, face several limitations that restrict their effectiveness. First, GCNs rely on fixed activation functions and trainable weights, restricting their ability to dynamically adapt to variations in human poses. Second, GCNs primarily aggregate information from one-hop neighbors, limiting their receptive field and making it challenging to capture long-range dependencies that are crucial for handling occlusions and depth ambiguities in 3D pose estimation. Third, GCNs exhibit spectral bias due to their use of MLPs for feature transformation.

\medskip\noindent\textbf{Adaptive Graph KAN.}\quad Inspired by our fixed point iterative solution given by \cref{Eq:FP} and to overcome the aforementioned limitations of GCN-based methods, we introduce an adaptive graph KAN (PoseKAN) model, which employs learnable function-based transformations in lieu of fixed activation functions and trainable weight matrices. Specifically, we define the layer-wise update rule of PoseKAN for node feature propagation as follows:
\begin{equation}
\bm{H}^{(\ell+1)}=\text{KAN}^{(\ell)}\Bigl(((1-s)\hat{\bm{A}}+s\hat{\bm{A}}^{2})\bm{H}^{(\ell)}
+\bm{X}\Bigr),
\label{Eq:IS}
\end{equation}
where $\ell\in\{0,\dots,L-1\}$, $L$ is the total number of layers, $\text{KAN}^{(\ell)}$ is a learnable single KAN layer, $s\in (0,1)$ is a positive scaling parameter that adjusts the contribution of immediate and second-order neighbors, and $\bm{H}^{(\ell)}\in\mathbb{R}^{J\times F_{\ell}}$ is the input feature matrix at the $\ell$-th layer with embedding dimension $F_{\ell}$. The input of the first layer is $\bm{H}^{(0)}=\bm{X}$. Note that unlike GCN, which uses fixed activation functions, PoseKAN learns its own activation functions dynamically, making it more expressive.

The PoseKAN update rule can be decomposed into two main operations: feature propagation and feature embedding. Feature propagation is given by
\begin{equation}
\bm{G}^{(\ell)}=\bm{P}\bm{H}^{(\ell)}+\bm{X},
\end{equation}
where $\bm{P}=(1-s)\hat{\bm{A}}+s\hat{\bm{A}}^{2}$ is the propagation matrix defined as a weighted combination of the normalized adjacency matrix and its square, ensuring that multi-hop dependencies are captured. The addition of a residual connection ensures that information from the initial feature matrix is preserved throughout the layers. The parameter $s$ plays a crucial role in controlling the balance between local and higher-order information. After propagation, the feature embedding step refines the node representations by applying a KAN layer as follows:
\begin{equation}
\bm{H}^{(\ell+1)}=\text{KAN}^{(\ell)}(\bm{G}^{(\ell)}),
\end{equation}
where each edge in the graph is associated with a learnable univariate function via $\text{KAN}^{(\ell)}$, providing greater flexibility in feature adaptation.

While both PoseKAN and its GCN counterpart use learnable transformations, GCN employs learnable weight matrices for feature transformations, but these transformations are still linear mappings applied at each node. PoseKAN, on the other hand, employs learnable activation functions on edges rather than nodes. This not only allows for greater flexibility and adaptability, but also leads to improved expressiveness and mitigates spectral bias.

\begin{figure}[!htb]
\centering
\includegraphics[width=1\linewidth]{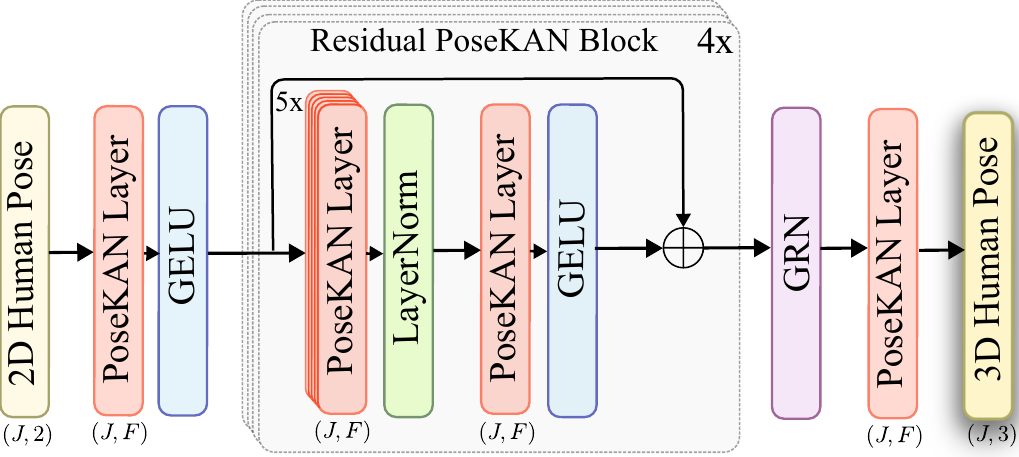}
\caption{\textbf{Overview of Model Architecture}. The model takes 2D pose coordinates as input and produces 3D pose predictions as output, where $J$ is the number of joints and $F$ is the embedding dimension. The architecture consists of a start PoseKAN layer, four residual FG-GCN blocks, and an end PoseKAN layer. A global response normalization is also used.}
\label{Fig:Architecture}
\end{figure}

\medskip\noindent\textbf{Model Architecture.}\quad The overall framework of our PoseKAN model architecture is depicted in Figure~\ref{Fig:Architecture}. The input to the model consists of 2D keypoints, obtained via an off-the-shelf 2D pose detector~\cite{chen2018cascaded}. Unlike GCN-based methods that use fixed activation functions, PoseKAN employs learnable function-based transformations, enabling adaptive, expressive, and interpretable feature learning. The PoseKAN architecture consists of start and end PoseKAN layers, and four residual PoseKAN blocks. The PoseKAN layer is the core computational unit of the model. The start PoseKAN layer maps input pose representations into latent space for effective feature learning, while the end PoseKAN layer projects refined feature representations back to pose space for final 3D pose predictions. Each residual PoseKAN block is comprised of five PoseKAN layers that learn hierarchical pose features, a layer normalization for stabilized training, an additional PoseKAN layer followed by the Gaussian Error Linear Unit (GELU) nonlinearity, and a residual connection to retain the original information and prevent gradient vanishing. GELU effectively preserves input magnitudes by smoothly blending linear and nonlinear transformations, thereby enhancing adaptability and expressiveness. The end PoseKAN layer generates the 3D pose, and is preceded by global response normalization (GRN)~\cite{woo2023convnext} to ensure feature contrast before prediction.

\medskip\noindent\textbf{Model Prediction.}\quad The end PoseKAN layer serves as the output transformation stage, where the learned feature representations are projected back into the 3D pose space to generate the final predicted joint coordinates. The output of this layer consists of the final node embeddings, $\hat{\bm{y}}_{i}$, which is the predicted 3D pose coordinates of $i$-th joint. Global response normalization is applied before the output projection to ensure that the output feature magnitudes are well-calibrated, reducing noise, and improving generalization.

\medskip\noindent\textbf{Model Training.}\quad The parameters (i.e., univariate activation functions) of the PoseKAN model are learned by minimizing the following loss function:
\begin{equation}
\mathcal{L} = \frac{1}{N} \Bigl[ (1 - \alpha) \sum_{i=1}^{N} \| \bm{y}_i - \hat{\bm{y}}_i \|_2^2 + \alpha \sum_{i=1}^{N} \| \bm{y}_i - \hat{\bm{y}}_i \|_1 \Bigr],
\end{equation}
which is a weighted combination of the mean squared error (MSE) and mean absolute error (MAE) between the true 3D joint coordinates $\bm{y}_i$ and the predicted 3D joint coordinates $\hat{\bm{y}}_i$ across $N$ training joints. The weight $\alpha\in [0,1]$ controls the contribution of each error term. Our proposed loss function $\mathcal{L}$ draws inspiration from the penalty function used in the elastic net regression model~\cite{Zou2005Elastic}, which is a weighted combination of lasso and ridge regularization.

\medskip\noindent\textbf{Model Complexity Analysis.}\quad For simplicity, we assume the embedding dimensions are the same across all layers, i.e., $F_{\ell}=F$ for all $\ell$. The computational cost of multiplying the propagation matrix $\bm{P}$ with the embedding $\bm{H}^{(\ell)}$ is $\mathcal{O}(\Vert\hat{\bm{A}}\Vert_{0}F)$ in time, where $\Vert\hat{\bm{A}}\Vert_{0}$ represents the number of nonzero entries in the sparse normalized adjacency matrix $\hat{\bm{A}}$, effectively corresponding to the number of edges in the graph. This term quantifies the complexity of message passing in the graph structure. Applying a KAN layer incurs a computational cost of $\mathcal{O}(GF^2)$, where $G$ is the grid size. Consequently, the overall \textit{time complexity} of an $L$-layer PoseKAN is $\mathcal{O}(L\Vert\hat{\bm{A}}\Vert_{0}F+LGF^2)$, where the first term corresponds to feature propagation, while the second term arises from KAN-based feature transformations. Regarding \textit{memory complexity}, an $L$-layer PoseKAN requires $\mathcal{O}(LJF+2^{k}GLF^2)$ in memory, where $\mathcal{O}(LJF)$ accounts for storing feature embeddings across all layers and $\mathcal{O}(2^{k}GLF^2)$ represents memory usage in $L$ KAN layers, with $2^{k}$ stemming from the recursive computation of order $k$ splines used in function approximation. By comparison, a standard GCN has a relatively lower memory complexity of $\mathcal{O}(LJF+LF^2)$, as it lacks the function-based transformations introduced by KAN layers. However, since $k$ and $G$ are typically small in practical implementations, the additional cost remains manageable. A key computational optimization in PoseKAN is that it avoids explicitly computing the square of the normalized adjacency matrix. Instead, we apply a right-to-left multiplication strategy, where $\hat{\bm{A}}$ is first multiplied with the feature embedding before the second multiplication is performed. This optimization eliminates unnecessary matrix exponentiation, significantly reducing computational overhead while preserving the benefits of multi-hop feature propagation. As a result, PoseKAN achieves a strong balance between computational efficiency and enhanced model expressiveness.

\section{Experiments}
\subsection{Experimental Setup}
\noindent\textbf{Datasets.}\quad We conduct experimental evaluations on two standard datasets: Human3.6M~\cite{ionescu2013human3} and MPI-INF-3DHP~\cite{Dushyant:2017}. We follow standard protocols~\cite{martinez2017simple,zhao2019semantic} for data preprocessing and splitting on these benchmark datasets.

\medskip\noindent\textbf{Evaluation Protocols and Metrics.}\quad We employ two standard evaluation protocols for training and testing on Human 3.6M designated as Protocol \#1 and Protocol \#2 ~\cite{martinez2017simple}, with associated metrics mean per-joint position error (MPJPE) and Procrustes-aligned mean per-joint position error (PA-MPJPE), respectively. For MPI-INF-3DHP, we use the Area Under Curve (AUC) and Percentage of Correct Keypoint (PCK) as assessment metrics.

\medskip\noindent\textbf{Baseline Methods.}\quad We compare the performance of our model with several state-of-the-art methods for 3D pose estimation, including semantic GCN (SemGCN)~\cite{zhao2019semantic}, High-order GCN~\cite{zou2020high}, Compositional GCN (CompGCN)~\cite{zou2021compositional}, Higher-Order Implicit Fairing Network (HOIF-Net)~\cite{quan2021higher}, Multi-hop Modulated GCN (MM-GCN)~\cite{lee2022multi}, Group GCN~\cite{zhang2022group}, Modulated GCN~\cite{zou2021modulated}, Flexible GCN (Flex-GCN)~\cite{Shah2024FlexGCN}, and GraphMLP~\cite{li2025graphmlp}.

\medskip\noindent\textbf{Implementation Details.}\quad Our model is implemented in PyTorch and all experiments are conducted on a single NVIDIA RTX A4500 GPU with 20GB of memory. We employ the AMSGrad optimizer for training, running for 30 epochs on both 2D ground truth and 2D pose detections~\cite{chen2018cascaded}. The initial learning rate is set to 0.001, with a decay factor of 0.99 every four epochs. The batch size and embedding dimension are set to 64 and $F=240$, respectively. The scaling parameter $s=0.2$ and weighting factor $\alpha=0.03$ are determined via grid search. To prevent overfitting, we apply dropout with a factor of 0.2 after each PoseKAN layer. The spline order and grid size are set to 3 and 5, respectively.

\begin{table*}[!htb]
\caption{Comparison of our model and baseline methods in terms of Mean Per Joint Position Error (MPJPE) in millimeters, computed between the ground truth and estimated poses on the Human3.6M dataset under Protocol \#1. The last column displays the average errors. The best results are shown in \textbf{bold}, and the second best results are \underline{underlined}.}
\small
\setlength{\tabcolsep}{2.8pt}
\smallskip
\centering
\begin{tabular}{l*{17}{c}}
\toprule
& \multicolumn{15}{c}{Action}\\
\cmidrule(lr){2-16}
Method & Dire. & Disc. &  Eat & Greet & Phone & Photo &  Pose & Purch. & Sit & SitD. & Smoke & Wait & WalkD. & Walk & WalkT. & Avg.\\
\midrule
Ordinal Depth~\cite{pavlakos2018ordinal} & 48.5& 54.4& 54.4& 52.0 &59.4 &65.3 &49.9& 52.9& 65.8 &71.1& 56.6& 52.9& 60.9& 44.7& 47.8& 56.2\\
MultiPoseNet~\cite{sharma2019monocular} & 48.6 &54.5& 54.2& 55.7& 62.2& 72.0& 50.5& 54.3& 70.0& 78.3 &58.1& 55.4& 61.4& 45.2& 49.7& 58.0\\
SemGCN~\cite{zhao2019semantic} & 47.3& 60.7& 51.4 &60.5& 61.1& \textbf{49.9} & 47.3 & 68.1 &86.2& \textbf{55.0}& 67.8& 61.0& \textbf{42.1} & 60.6& 45.3& 57.6\\
WSGN~\cite{li2020weakly} & 62.0 & 69.7 & 64.3 & 73.6 & 75.1 & 84.8 & 68.7 & 75.0 & 81.2 & 104.3 & 70.2 & 72.0 & 75.0 & 67.0 & 69.0 & 73.9\\
PoseGraphNet~\cite{banik20223d} & 51.0 & 55.3 & 54.0 & 54.6 & 62.4 & 76.0 & 51.6 & 52.7 & 79.3 & 87.1 & 58.4 & 56.0 & 61.8 & 48.1 & 44.1 & 59.5\\
PGDA~\cite{xu2021monocular} & 47.1 & 52.8 & 54.2 & 54.9 & 63.8 & 72.5 & 51.7 & 54.3 & 70.9 & 85.0 & 58.7 & 54.9 & 59.7 & 43.8 & 47.1 & 58.1\\
High-order GCN~\cite{zou2020high} & 49.0& 54.5& 52.3& 53.6& 59.2 &71.6& 49.6& 49.8 &66.0 &75.5 &55.1 &53.8& 58.5& 40.9 & 45.4 &55.6\\
HOIF-Net~\cite{quan2021higher} & 47.0 & 53.7 & 50.9 & 52.4 & 57.8 & 71.3 & 50.2 & 49.1 & 63.5 & 76.3 & 54.1 & 51.6 & 56.5 & 41.7 & 45.3 & 54.8\\
CompGCN~\cite{zou2021compositional} & 48.4 & 53.6 & 49.6 & 53.6 & 57.3 & 70.6 & 51.8 & 50.7 & 62.8 & 74.1 & 54.1 & 52.6 & 58.2 & 41.5 & 45.0 & 54.9\\
Modulated GCN~\cite{zou2021modulated} & 45.4 & \underline{49.2} & 45.7 & 49.4 & \underline{50.4} & 58.2 & 47.9 & 46.0 & 57.5 & 63.0 & 49.7 & 46.6 & 52.2 & 38.9 & 40.8 & 49.4\\
MM-GCN~\cite{lee2022multi} & 46.8 & 51.4 & 46.7 & 51.4 & 52.5 & 59.7 & 50.4 & 48.1 & 58.0 & 67.7 & 51.5 & 48.6 & 54.9 & 40.5 & 42.2 & 51.7\\
Group GCN~\cite{zhang2022group} & 45.0 & 50.9 & 49.0 & 49.8 & 52.2 & 60.9 & 49.1 & 46.8 & 61.2 & 70.2 & 51.8 & 48.6 & 54.6 & 39.6 & 41.2 & 51.6\\
GraphMLP~\cite{li2025graphmlp} & \underline{43.7} & 49.3 & \underline{45.5} & \underline{47.9} & 50.5 & 56.0 & \underline{46.3} & \underline{44.1} & \underline{55.9} & \underline{59.0} & \underline{48.4} & \underline{45.7} & 51.2 & \textbf{37.1} & \underline{39.1} & \underline{48.0} \\
\midrule
PoseKAN (ours) & \textbf{40.9} & \textbf{45.6} & \textbf{44.4} & \textbf{47.4} &\textbf{48.4} & \underline{52.8} & \textbf{44.1} & \textbf{41.6} & \textbf{54.6} & 63.8  &\textbf{46.1} & \textbf{45.1} & 49.0 & \underline{38.1} & \textbf{39.0} & \textbf{46.7} \\
\bottomrule
\end{tabular}
\label{Tab:MPJPE_Result}
\end{table*}

\begin{table*}[!htb]
\caption{Comparison of our model and baseline methods in terms of Procrustes-aligned Mean Per Joint Position Error (PA-MPJPE), computed between the ground truth and estimated poses on the Human3.6M dataset under Protocol \#2.}
\small
\setlength{\tabcolsep}{2.8pt}
\smallskip
\centering
\begin{tabular}{l*{17}{c}}
\toprule
& \multicolumn{15}{c}{Action}\\
\cmidrule(lr){2-16}
Method & Dire. & Disc. &  Eat & Greet & Phone & Photo &  Pose & Purch. & Sit & SitD. & Smoke & Wait & WalkD. & Walk & WalkT. & Avg.\\
\midrule
p-LSTMs~\cite{lee2018propagating}  & 38.0 & 39.3 & 46.3 & 44.4 & 49.0 & 55.1 & 40.2 & 41.1 & 53.2 & 68.9 & 51.0 & 39.1 & \textbf{33.9} & 56.4 & 38.5 & 46.2 \\
WSGN~\cite{li2020weakly} & 38.5 & 41.7 & 39.6 & 45.2 & 45.8 & 46.5 & 37.8 & 42.7 & 52.4 & 62.9 & 45.3 & 40.9 & 45.3 & 38.6 & 38.4 & 44.3\\
PoseGraphNet~\cite{banik20223d} & 38.4 & 43.1 & 42.9 & 44.0 & 47.8 & 56.0 & 39.3 & 39.8 & 61.8 & 67.1 & 46.1 & 43.4 & 48.4 & 40.7 & 35.1 & 46.4\\
PGDA~\cite{xu2021monocular} & 36.7 & 39.5 & 41.5 & 42.6 & 46.9 & 53.5 & 38.2 & 36.5 & 52.1 & 61.5 & 45.0 & 42.7 & 45.2 & 35.3 & 40.2 & 43.8\\
High-order GCN~\cite{zou2020high} &38.6 &42.8& 41.8 &43.4 &44.6& 52.9& 37.5& 38.6 &53.3 &60.0& 44.4& 40.9& 46.9 &32.2 &37.9 &43.7\\
HOIF-Net~\cite{quan2021higher} & 36.9 & 42.1 & 40.3 & 42.1 & 43.7 & 52.7 & 37.9 & 37.7 & 51.5 & 60.3 & 43.9 & 39.4 & 45.4 & 31.9 & 37.8 & 42.9\\
CompGCN~\cite{zou2021compositional} & 38.4 & 41.1 & 40.6 & 42.8 & 43.5 & 51.6 & 39.5 & 37.6 & 49.7 & 58.1 & 43.2 & 39.2 & 45.2 & 32.8 & 38.1 & 42.8\\
Modulated GCN~\cite{zou2021modulated} & 35.7 & 38.6 & \underline{36.3} & 40.5 & \underline{39.2} & 44.5 & 37.0 & 35.4 & 46.4 & \underline{51.2} & 40.5 & \underline{35.6} & 41.7 & 30.7 & 33.9 & 39.1\\
MM-GCN~\cite{lee2022multi} & 35.7 & 39.6 & 37.3 & 41.4 & 40.0 & 44.9 & 37.6 & 36.1 & 46.5 & 54.1 & 40.9 & 36.4 & 42.8 & 31.7 & 34.7 & 40.3\\
Group GCN~\cite{zhang2022group} & 35.3 & 39.3 & 38.4 & 40.8 & 41.4 & 45.7 & 36.9 & 35.1 & 48.9 & 55.2 & 41.2 & 36.3 & 42.6 & 30.9 & 33.7 & 40.1\\
GraphMLP~\cite{li2025graphmlp} & \underline{35.1} & \underline{38.2} & 36.5 & \textbf{39.8} & 39.8 & \underline{43.5} & \underline{35.7} & \underline{34.0} & \underline{45.6} & \textbf{47.6} & \underline{39.8} & \textbf{35.1} & \underline{41.1} & \textbf{30.0} & \underline{33.4} & \underline{38.4} \\
\midrule
PoseKAN (ours) & \textbf{34.0} & \textbf{37.2} & \textbf{36.0} & \underline{40.0} &\textbf{39.0} & \textbf{42.9} & \textbf{35.1} & \textbf{33.3} & \textbf{45.4} & 54.5 & \textbf{38.3} & 36.0 & \textbf{40.5} & \underline{30.1} & \textbf{32.4} &\textbf{38.3}\\
\bottomrule
\end{tabular}
\label{Tab:PA_MPJPE_Result}
\end{table*}

\subsection{Results and Analysis}
\noindent\textbf{Quantitative Results on Human3.6M.}\quad In Tables~\ref{Tab:MPJPE_Result} and~\ref{Tab:PA_MPJPE_Result}, we report performance comparison of our PoseKAN model and strong baselines for 3D pose estimation on Human3.6M using the detected 2D pose as input. In both tables, we present the results of all 15 actions, as well as the average performance. Table~\ref{Tab:MPJPE_Result} demonstrates that under Protocol \#1, PoseKAN consistently outperforms state-of-the-art models across multiple action categories. Compared to GraphMLP, the strongest competing baseline, PoseKAN achieves a relative error reduction of 2.7\%. Against Modulated GCN, our model achieves a 5.47\% reduction in error and performs better in 14 out of 15 action categories, emphasizing the superiority of learnable function-based transformations over pre-defined activation functions in traditional GCNs. Furthermore, PoseKAN significantly outperforms High-Order GCN, achieving a relative error reduction of 15.99\%, showcasing its ability to capture multi-hop dependencies more effectively without relying on fixed high-order convolutions. Our model not only excels in standard poses but also demonstrates superior generalization on challenging actions involving significant self-occlusions, such as Eating, Sitting, and Smoking. PoseKAN effectively mitigates these challenges through multi-hop information aggregation, enabling the model to recover occluded joint positions more accurately than prior GCN-based methods.

Under Protocol \#2, Table~\ref{Tab:PA_MPJPE_Result} demonstrates that PoseKAN achieves the lowest PA-MPJPE of 38.3mm, outperforming all baseline models. When compared to GraphMLP, PoseKAN achieves a relative error reduction of 0.26\%, indicating a modest yet consistent improvement over the best competing baseline. Moreover, PoseKAN outperforms GraphMLP in 11 out of 15 action categories, reinforcing its ability to generalize across a diverse range of human activities. Similarly, PoseKAN reduces the error by 2.04\% compared to Modulated GCN and achieves better results in 13 out of 15 actions. Notably, PoseKAN demonstrates superior performance in occlusion-prone actions, outperforming Modulated GCN by 1.23\% in Greeting, 2.15\% in Sitting, and 5.43\% in Smoking, highlighting its robustness in handling self-occlusions and complex body poses. Furthermore, PoseKAN surpasses Modulated GCN on the \ challenging Photo action, achieving a relative error reduction of 3.59\%, indicating better feature learning in scenarios with complex backgrounds and fine-grained pose details. Most significantly, PoseKAN outperforms High-Order GCN with a substantial relative error reduction of 12.35\%, achieving lower errors across all action categories.

\begin{figure*}[!htb]
\centering
\includegraphics[width=.86\linewidth]{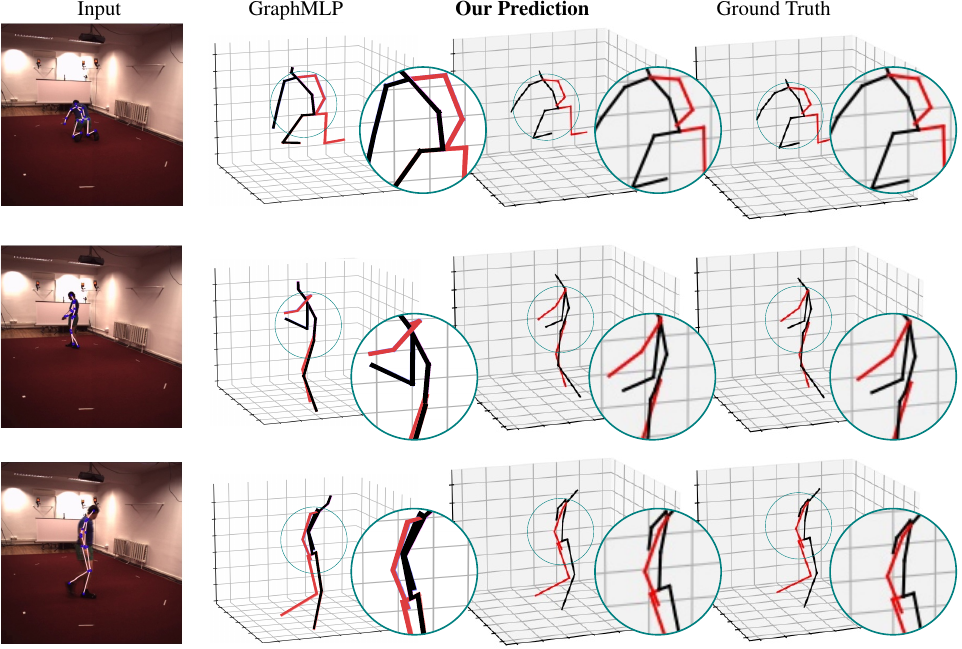}
\caption{Visual comparison between PoseKAN and GraphMLP on sample actions from the Human3.6M dataset.}
\label{Fig:Qualitative}
\end{figure*}

\medskip\noindent\textbf{Qualitative Results.}\quad Figure~\ref{Fig:Qualitative} presents qualitative visualizations of PoseKAN's predictions on the Human3.6M dataset across various action categories, illustrating the model's effectiveness in accurately estimating 3D human poses from monocular 2D inputs. The predicted 3D poses exhibit a high degree of alignment with the ground truth, demonstrating the model's ability to capture spatial dependencies and maintain structural consistency even in complex motion scenarios. The precise joint placements and natural articulation of limbs in our predictions further validate PoseKAN's ability to mitigate depth ambiguities and occlusions, which are inherent challenges in 2D-to-3D pose estimation. A comparative analysis with GraphMLP further highlights PoseKAN's advantages in handling challenging cases involving self-occlusion, where certain body parts obscure others, making pose reconstruction significantly harder. Unlike GraphMLP, which occasionally produces unnatural joint positions or misaligned skeletal structures, PoseKAN generates pose estimations that closely resemble the ground truth pose, capturing fine-grained joint interactions more effectively.

\begin{table}[!htb]
\caption{Results on MPI-INF-3DHP.}
\setlength{\tabcolsep}{5pt}
\smallskip
\centering
\begin{tabular}{lcc}
\toprule
Method & PCK $(\uparrow)$ & AUC $(\uparrow)$\\
\midrule
Ordinal Depth~\cite{pavlakos2018ordinal} & 71.9 & 35.3 \\
Pose Attribute~\cite{wang2019not} & 71.9 & 35.8 \\
HOIF-Net~\cite{quan2021higher} & 72.8 & 36.5 \\
LCN~\cite{ci2019optimizing} & 74.0 & 36.7 \\
HEMlets~\cite{zhou2019hemlets} & 75.3 & 38.0 \\
SRNet~\cite{zeng2020srnet} & 77.6 & 43.8 \\
Weight Unsharing~\cite{liu2020comprehensive} & 79.3 & 47.6 \\
CompGCN~\cite{zou2021compositional} & 79.3 & 45.9 \\
GraphSH~\cite{xu2021graph} & 80.1 & 45.8 \\
HCSF~\cite{zeng2021learning} & 82.1 & 46.2 \\
MM-GCN~\cite{lee2022multi} & 81.6 & 50.3\\
Group GCN~\cite{zhang2022group} & 81.1 & 49.9\\
Flex-GCN~\cite{Shah2024FlexGCN} & 85.2 & 51.8 \\
ICFNet~\cite{Wang2025ICFNet} & \underline{85.6} & \textbf{54.3} \\
\midrule
PoseKAN (ours) & \textbf{86.0} & \underline{52.9}\\
\bottomrule
\end{tabular}
\label{Tab:mpi_3dhp_inf}
\end{table}

\medskip\noindent\textbf{Cross-Dataset Results on MPI-INF-3DHP.}\quad In Table~\ref{Tab:mpi_3dhp_inf}, we evaluate the generalization ability of our method by comparing it against strong baselines. Our model is trained on Human3.6M and evaluated on the MPI-INF-3DHP dataset. Results demonstrate that our approach achieves the highest PCK and second best AUC scores, consistently outperforming the baseline methods across various indoor and outdoor scenes. Compared to ICFNet, the best performing baseline, our model shows a relative improvement of 0.5\% in terms of the PCK  metric. While ICFNet achieves a slightly higher AUC score, PoseKAN remains highly competitive.

\medskip\noindent\noindent\textbf{Model Efficiency.}\quad PoseKAN achieves superior pose estimation performance while being significantly more efficient, utilizing only 5.72M parameters, much lower than GraphMLP's 9.49M parameters. Although our model incurs a slight computational overhead compared to GCN-based methods, it significantly improves performance.

\subsection{Ablation Study}
\noindent\textbf{Effect of Residual Connection.}\quad  We analyze the impact of the initial residual connection (IRC) within the layer-wise propagation rule on our model's performance. The results, summarized in Table~\ref{Tab:skipConnection}, indicate that incorporating IRC leads to notable performance improvements, reducing MPJPE and PA-MPJPE by 1.65\% and 1.49\%, respectively. By preserving and reinforcing the initial node features across layers, IRC not only stabilizes the learning process but also mitigates information loss, allowing for more reliable pose predictions.

\begin{table}[!htb]
\caption{Effect of initial residual connection (IRC) on model performance.}
\setlength{\tabcolsep}{2.5pt}
\smallskip
\centering
\begin{tabular}{l*{7}{c}}
\toprule
Method & MPJPE $(\downarrow)$ & PA-MPJPE $(\downarrow)$ \\
\midrule
Without IRC & 34.44 & 28.79 \\
With IRC & \textbf{33.51} & \textbf{28.01}\\
\bottomrule
\end{tabular}
\label{Tab:skipConnection}
\end{table}

\begin{figure*}[!htb]
\begin{minipage}[b]{.6\textwidth}
\centering
\setlength\tabcolsep{2pt} 
\begin{tabular}{cc}
\includegraphics[width=2in]{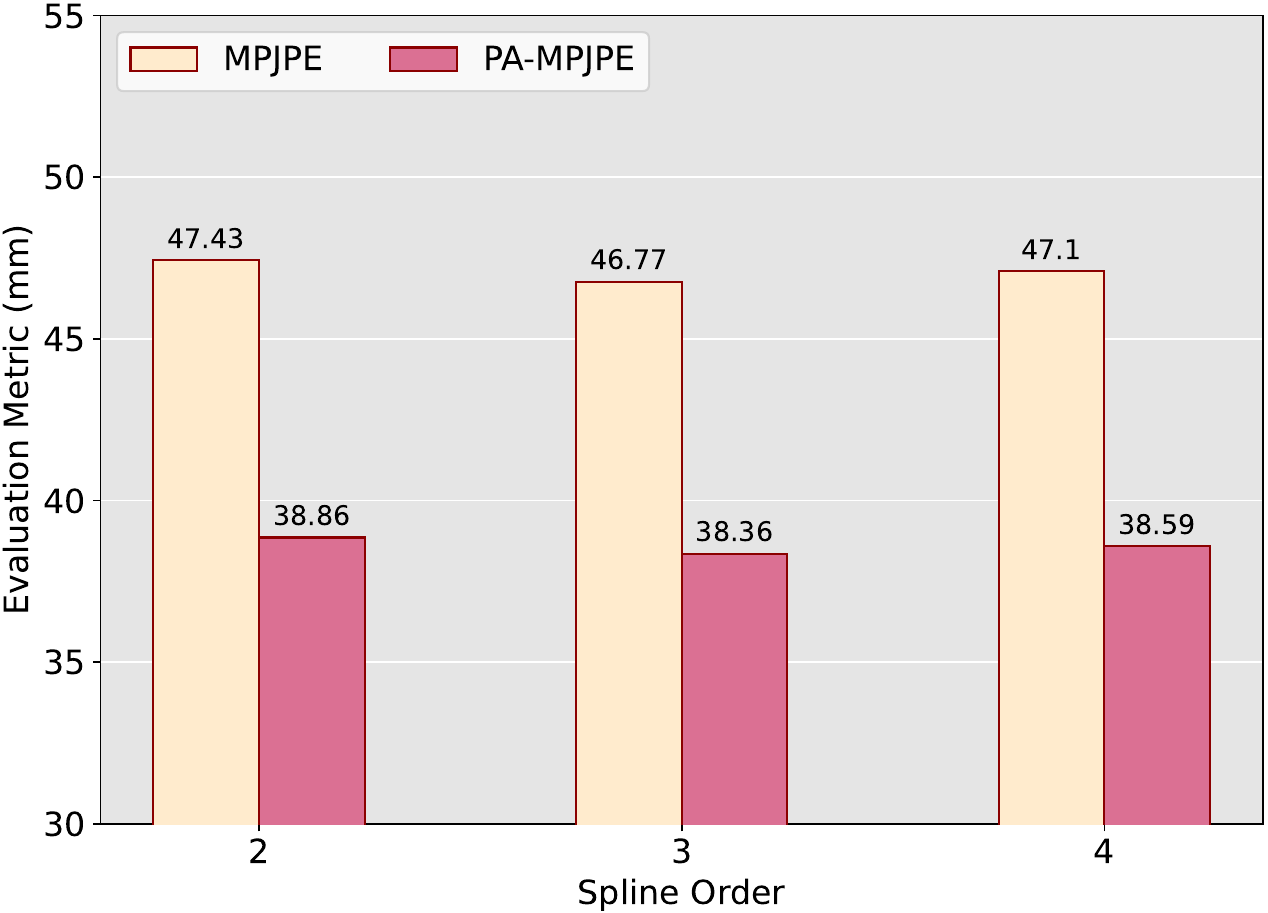} & \includegraphics[width=2in]{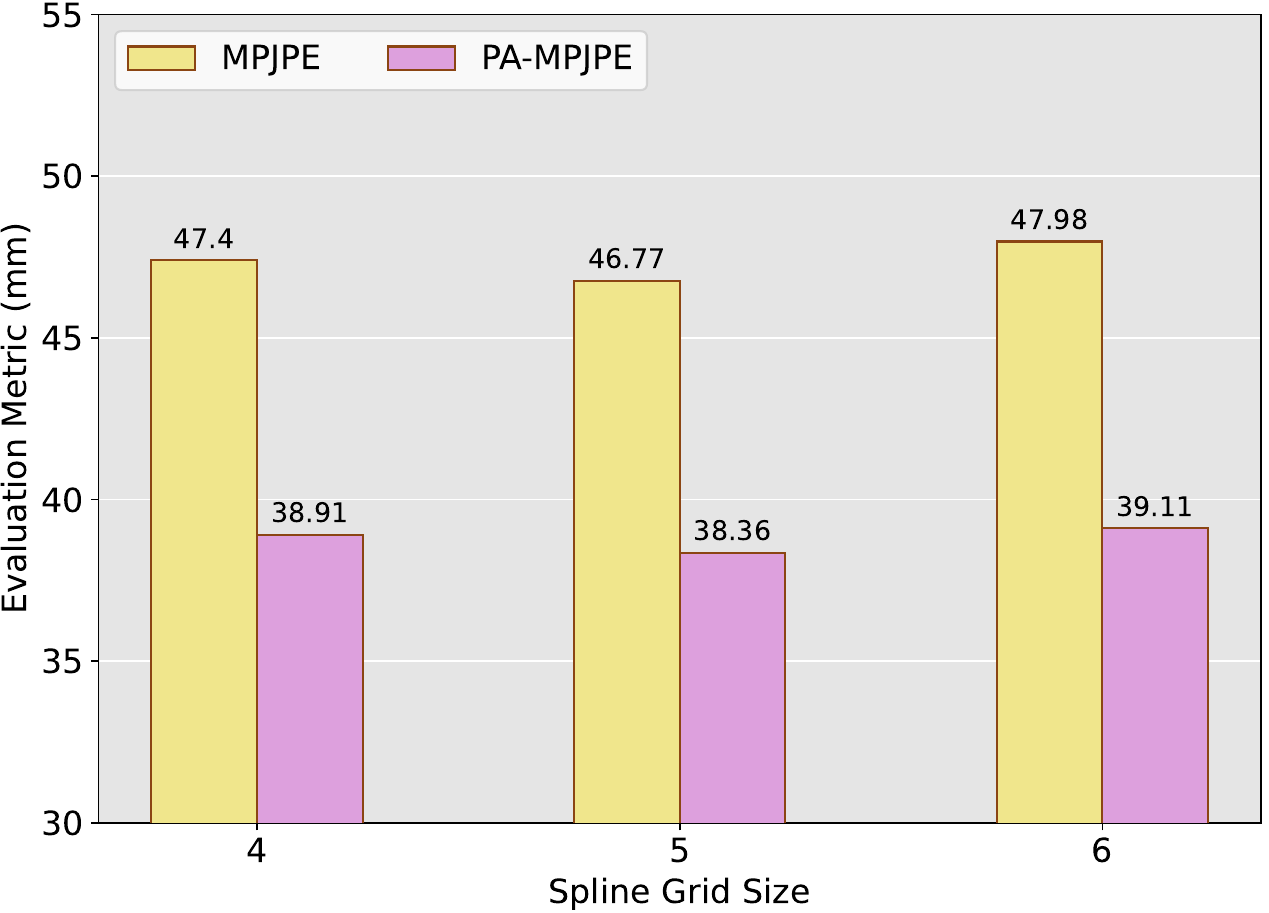}
\end{tabular}
\caption{Effect of spline order and grid size.}
\label{Fig:OrderGrid}
\end{minipage}
\begin{minipage}[b]{.4\textwidth}
\centering
\includegraphics[width=2.5in]{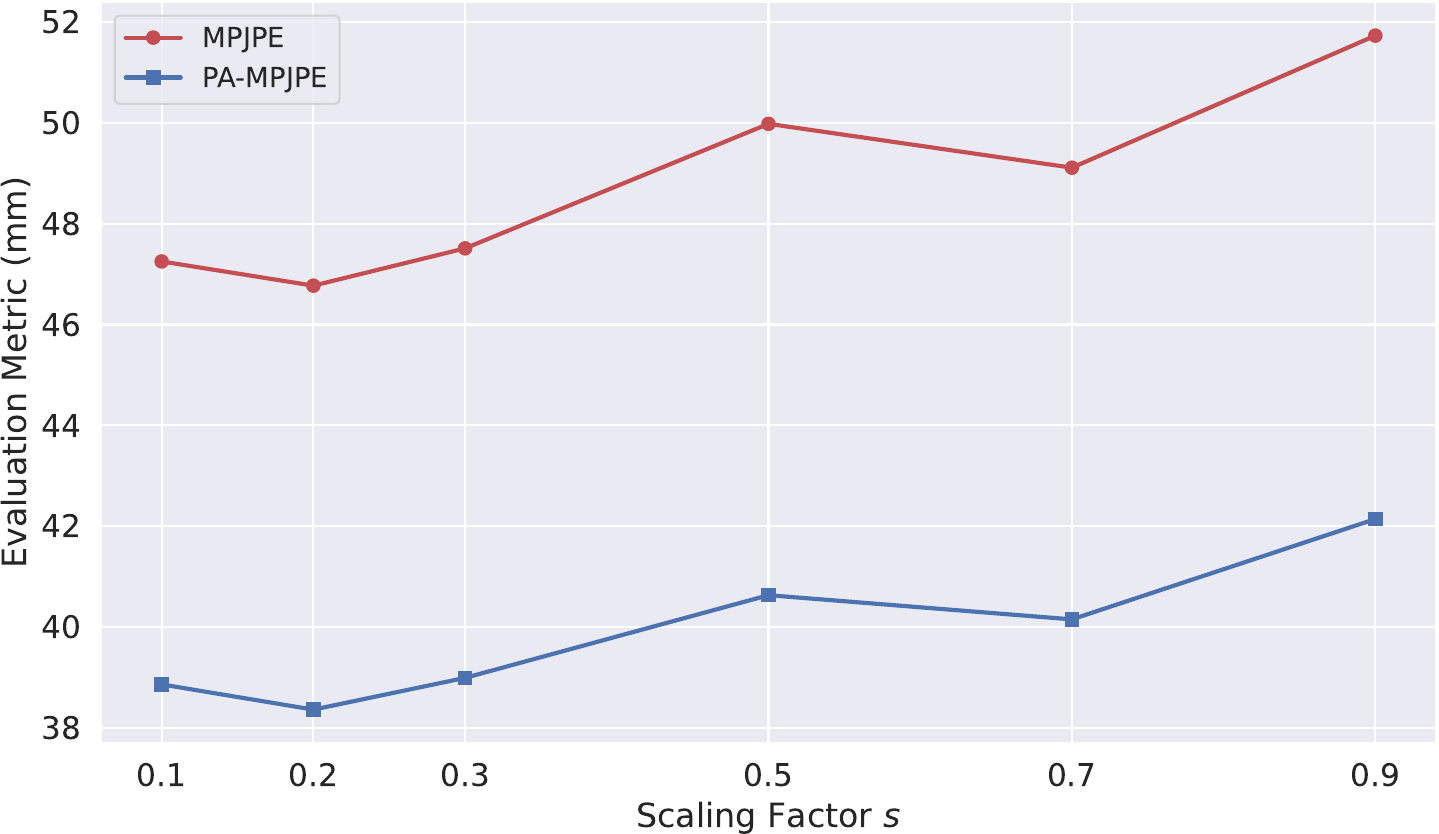}
\caption{Effect of scaling factor.}
\label{Fig:FigureHyperparameter}
\end{minipage}
\end{figure*}

\smallskip\noindent\textbf{Effect of Spline Order and Grid Size.} \quad In Figure~\ref{Fig:OrderGrid} (left),  we observe that the best performance is achieved at a spline order of 3. Increasing the spline order beyond this point does not lead to further improvements and may introduce unnecessary complexity. Similarly, Figure~\ref{Fig:OrderGrid} (right) shows that the best results are obtained at a grid size of 5. While a sufficiently high spline order and grid size allow for more adaptive function approximations, excessively large values lead to increased computational costs.

\smallskip\noindent\textbf{Effect of Scaling Factor.} \quad By varying the scaling factor $s$ from 0 and 1, we control how much weight is given to local versus global dependencies within the graph structure. This ability to adjust the influence of distant nodes is particularly valuable when learning graph representations that need to capture global patterns and long-range dependencies, which are essential for tasks like 3D human pose estimation. Figure~\ref{Fig:FigureHyperparameter} illustrates that smaller scaling parameters often lead to better results, with our model achieving the lowest error values at $s = 0.2$. This suggests that a moderate emphasis on local information, while still considering distant nodes, strikes the good balance for our model.

\section{Conclusion}
In this paper, we introduced an adaptive graph Kolmogorov-Arnold Network (PoseKAN), a novel approach for 3D human pose estimation that leverages learnable function-based transformations. Unlike GCN-based methods that rely on fixed activation functions, PoseKAN employs learnable univariate functions on graph edges, providing greater flexibility and expressiveness in modeling human skeletal structures. By incorporating multi-hop feature propagation, our model effectively captures both local and long-range dependencies, mitigating the challenges posed by occlusions and depth ambiguities in 3D pose estimation. Our extensive experiments on benchmark datasets demonstrate that PoseKAN outperforms state-of-the-art methods, while also achieving good performance in cross-dataset evaluations, proving its generalization ability to unseen data. For future work, we aim to extend PoseKAN to multi-person pose estimation tasks and explore its applicability to other graph-based learning problems, such as action recognition and anomaly detection.

{
    \small
    \bibliographystyle{ieeenat_fullname}
    \bibliography{references}
}

\clearpage
\setcounter{page}{1}
\maketitlesupplementary

\section{Multi-Layer Kolmogorov-Arnold Networks}
\begin{figure}[!htb]
\centering
\includegraphics[width=1\linewidth]{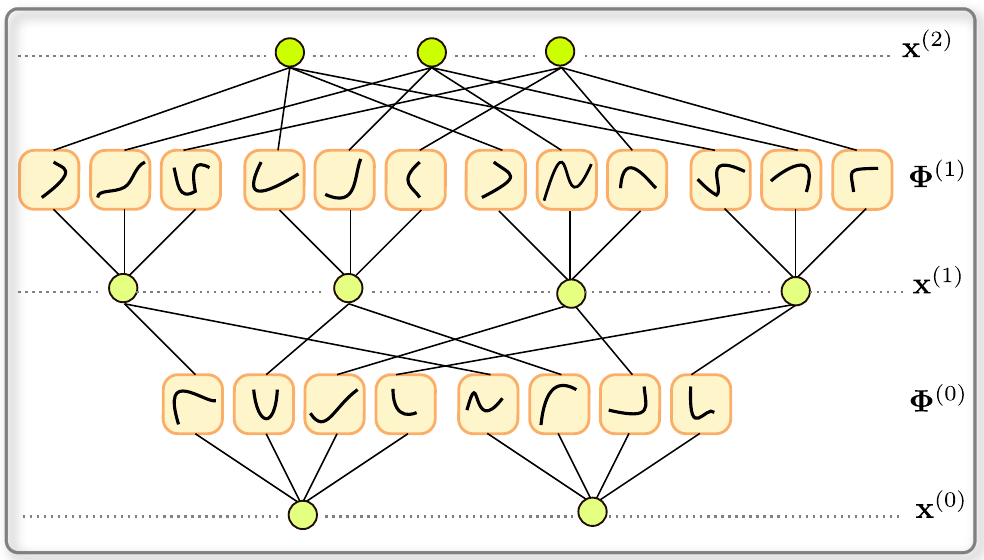}
\caption{Illustration of a two-layer KAN architecture.}
\label{Fig:KAN-architecture}
\end{figure}

\begin{align}
\bm{x}^{(\ell+1)} &= \text{KAN}^{(\ell)}(\bm{x}^{(\ell)}) \\
& =\underbrace{\begin{pmatrix}
    \phi_{1,1}^{(\ell)}(\cdot) & \dots  & \phi_{1,F_{\ell}}^{(\ell)}(\cdot)\\
    \vdots & \ddots & \vdots\\
    \phi_{F_{\ell+1},1}^{(\ell)}(\cdot) & \dots  & \phi_{F_{\ell+1},F_{\ell}}^{(\ell)}(\cdot)
    \end{pmatrix}}_{\bg{\Phi}^{(\ell)}} \bm{x}^{(\ell)}
\end{align}
where $\bg{\Phi}^{(\ell)}=(\phi_{q,p}^{(\ell)})$ is an $F_{\ell+1}\times F_{\ell}$ matrix of functions associated with $\text{KAN}^{(\ell)}$. Similar to MLPs, multiple KAN layers can be stacked to form a deep KAN, enabling the model to learn hierarchical representations and capture complex patterns more effectively. A deep KAN consists of $L$ layers, where each layer applies a transformation to the input using a matrix of learnable functions, as illustrated in Figure~\ref{Fig:KAN-architecture}.

\medskip

\section{Datasets}
\noindent\textbf{Human3.6M} is is a large-scale dataset containing images captured at 50Hz from four synchronized cameras, providing diverse viewpoints and configurations. It features 11 actors (6 male, 5 female) performing 15 indoor activities, including Directions, Discussion, Eating, Greeting, Phoning, Posing, Purchases, Sitting, Sitting Down, Smoking, Photo, Waiting, Walk Dog, Walking, and Walk Together. The dataset provides precise 3D joint coordinates obtained via motion capture, which are projected onto 2D poses using known camera parameters. Annotations include detailed 3D joint information for seven subjects, with a training set consisting of five actors (S1, S5, S6, S7, and S8) and a test set comprising two actors (S9 and S11). The dataset is balanced across activities and subjects to ensure consistency. Before model input, both 2D and 3D poses are normalized, and the hip joint is set as the root joint in 3D postures for zero-centering.

\medskip\noindent\textbf{MPI-INF-3DHP} is a diverse dataset that captures human motion in both confined indoor spaces and complex outdoor environments. It features video recordings of eight actors (4 male, 4 female) filmed from 14 different angles, each performing eight activity sets that encompass a broad spectrum of poses. These activities range from basic motions like walking and sitting to more dynamic and physically demanding actions, offering greater variety compared to the Human3.6M dataset. Each sequence lasts approximately one minute, with actors alternating between two outfits, one casual for daily scenarios and another plain-colored to facilitate augmentation. The dataset also provides ground-truth 3D joint position annotations.

\section{Performance and Model Size Comparison}
\begin{figure}[!htb]
\begin{center}
\includegraphics[width=1\linewidth]{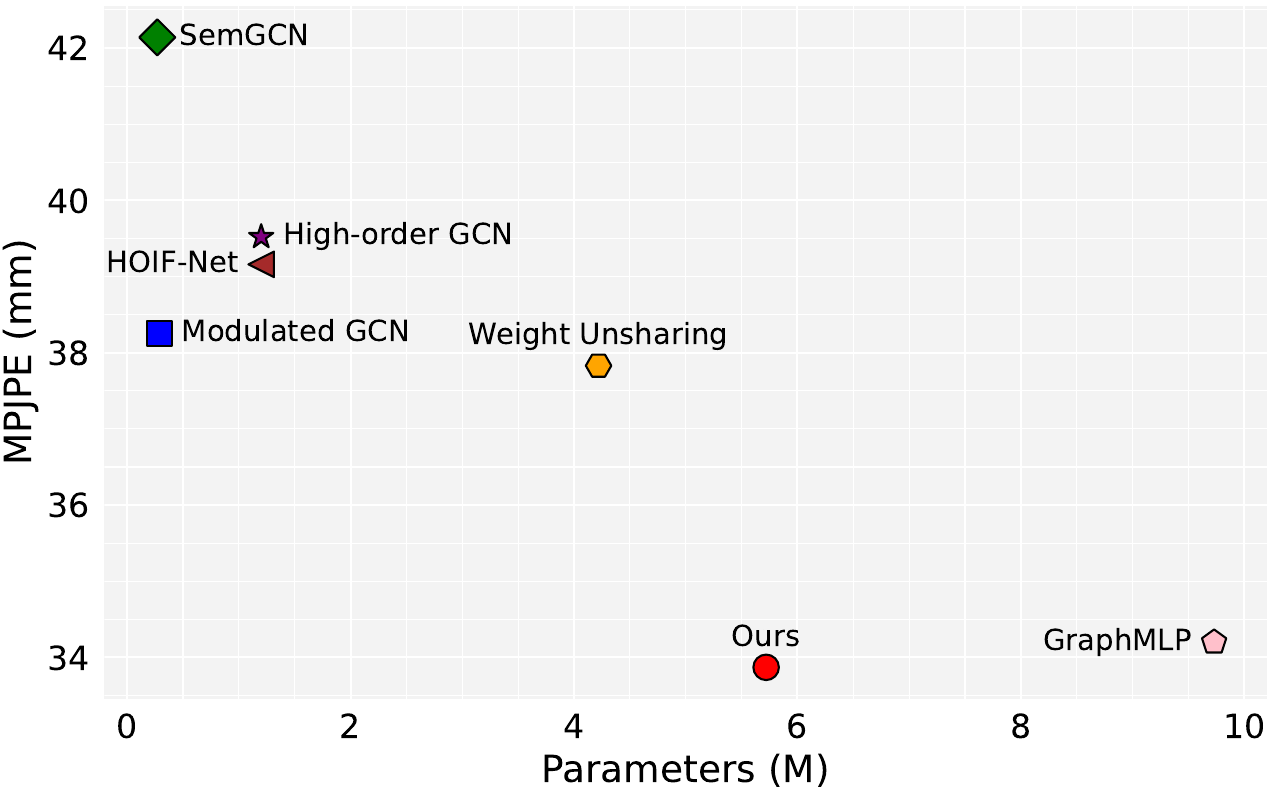}
\end{center}
\caption{Performance and model size comparison between our model and state-of-the-art  methods, including SemGCN~\cite{zhao2019semantic}, High-order GCN~\cite{zou2020high}, HOIF-Net~\cite{quan2021higher}, Modulated GCN~\cite{zou2021modulated} and GraphMLP~\cite{li2025graphmlp}. Lower Mean Per Joint Position Error (MPJPE) values indicate better performance. Evaluation is conducted on the Human3.6M dataset with ground truth 2D joints as input.}
\label{Fig:ModelSize}
\end{figure}

Although our model incurs a slight computational overhead compared to GCN-based methods, it significantly improves performance, as illustrated in Figure~\ref{Fig:ModelSize}, which compares the performance and model size of PoseKAN with state-of-the-art methods.

\section{Additional Experimental Results}
\medskip\noindent\textbf{Quantitative Results using Ground Truth.}\quad In Table~\ref{Tab:baselineComparison}, we report the performance comparison results under Protocol \#1 and Protocol \#2 for PoseKAN and several state-of-the-art baseline models, including SemGCN, High-Order GCN, HOIF-Net, Modulated GCN, Weight Unsharing, and GraphMLP, using ground truth 2D poses as input. These results highlight PoseKAN's superior performance across both evaluation protocols. Under Protocol \#1, PoseKAN achieves an MPJPE of 33.51mm, significantly outperforming all competing baselines. Specifically, compared to SemGCN, High-Order GCN, HOIF-Net, Modulated GCN, Weight Unsharing, and GraphMLP, our model reduces errors by 8.27mm, 5.65mm, 4.25mm, 4.38mm, 3.96mm, and 0.69mm, respectively. These improvements translate into relative error reductions of 19.62\%, 14.29\%, 11.14\%, 11.45\%, 10.46\%, and 2.01\%, demonstrating PoseKAN's ability to learn richer representations from skeletal graphs. Similarly, under Protocol \#2, PoseKAN achieves the lowest PA-MPJPE of 28.01mm, outperforming SemGCN, High-Order GCN, HOIF-Net, Modulated GCN, and Weight Unsharing. The relative error reductions amount to 15.41\%, 8.72\%, 4.64\%, 5.65\%, and 5.74\%, respectively

\medskip
\begin{table}[!htb]
\caption{Performance comparison of our model and other state-of-the-art methods using the 2D ground truth pose as input.}
\setlength{\tabcolsep}{3pt}
\smallskip
\centering
\begin{tabular}{lcc}
\toprule
Method & MPJPE $(\downarrow)$ & PA-MPJPE $(\downarrow)$ \\
\midrule
SemGCN~\cite{zhao2019semantic} & 42.14 & 33.53 \\
High-order GCN~\cite{zou2020high} & 39.52 & 31.07\\
HOIF-Net~\cite{quan2021higher} & 38.12 & \underline{29.74}\\
Modulated GCN~\cite{zou2021modulated} & 38.25 & 30.06\\
Weight Unsharing~\cite{liu2020comprehensive} & 37.83 & 30.09\\
GraphMLP~\cite{li2025graphmlp} & \underline{34.20}  &   -  \\
\midrule
PoseKAN (ours)  & \textbf{33.51} & \textbf{28.01} \\
\bottomrule
\end{tabular}
\label{Tab:baselineComparison}
\end{table}

\begin{figure}[!htb]
\centering
\includegraphics[width=1\linewidth]{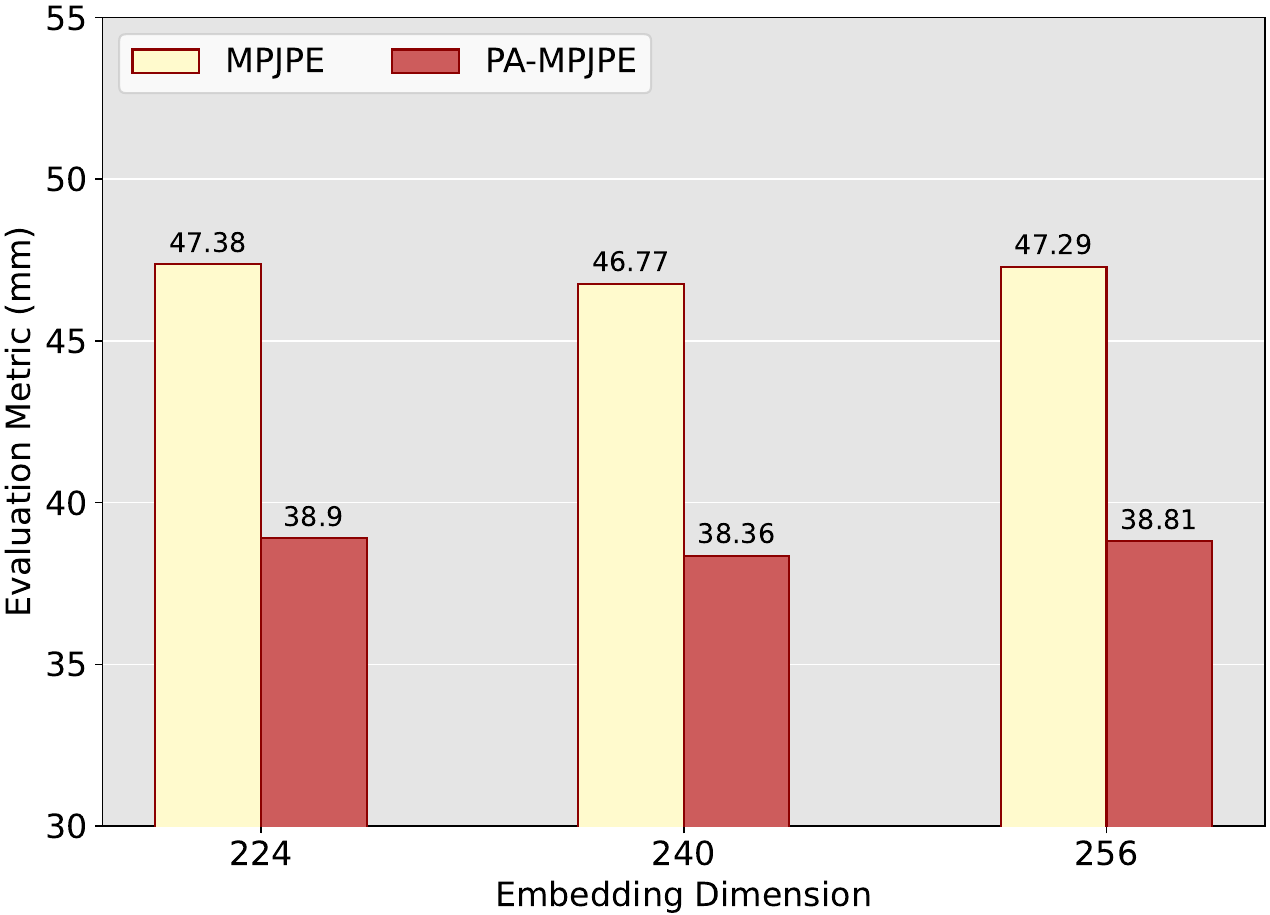}
\caption{Performance of our proposed PoseKAN model on the Human3.6M dataset using various embedding dimensions.}
\label{Fig:FigureBatchFilter}
\end{figure}

\medskip\noindent\textbf{Effect of Embedding Dimension.}\quad In Figure~\ref{Fig:FigureBatchFilter}, we analyze the impact of a varying embedding dimension on our model's performance. This hyperparameter, which determines the number of learnable parameters in each layer of the network, affects the model's capacity to capture complex patterns. Larger embedding dimensions enable the model to learn richer feature representations, improving its capacity to capture fine-grained motion details. However, increasing the embedding size excessively may lead to overfitting, where the model memorizes training data rather than generalizing to unseen samples. Smaller embedding dimensions reduce the number of parameters, making the model more lightweight and efficient, but may lead to underfitting and reduced prediction accuracy. Our analysis shows that the best performance is achieved using an embedding dimension of 240, providing a sufficient number of parameters to learn detailed features without overfitting.

\medskip

\section{Discussion}
The proposed PoseKAN framework presents several advantages over conventional GCN-based approaches for 3D human pose estimation. In this section, we highlight its merits in three key aspects: (1) Flexible and expressive feature learning, (2) Improved long-range dependency modeling, and (3) Reduced spectral bias for enhanced pose estimation.

\begin{itemize}
\item\textit{Flexible and Expressive Feature Learning}. Unlike standard GCN-based models that rely on fixed-weight transformations and predefined activation functions, PoseKAN leverages KANs, which introduce learnable activation functions on graph edges. These learnable functions provide greater expressiveness and data-driven adaptability.
\item\textit{Improved Long-Range Dependency Modeling}. Standard GCNs are inherently constrained by their reliance on one-hop message passing, which restricts the receptive field of each joint to its immediate neighbors. To overcome this limitation, PoseKAN incorporates a multi-hop propagation mechanism through the flexible propagation matrix with a scaling factor that balances information from immediate neighbors and higher-order dependencies, enabling joints that are not directly connected to exchange information. By modulating the adjacency matrix through a learnable modulation matrix, our model also learns new skeletal connections, further enhancing structure-aware learning.
\item\textit{Reduced Spectral Bias}. Since MLPs are core components in GCN-based methods, they exhibit spectral bias, where the model prefers learning low-frequency signals while struggling to capture high-frequency variations. By contrast, PoseKAN mitigates this issue by dynamically learning function-based transformations to capture fine-grained motion variations, allowing our model to learn both low- and high-frequency components more effectively:
\end{itemize}
\noindent\textbf{Limitations.}\quad While PoseKAN offers greater interpretability compared to GCN-based methods, visualizing how each learnable activation function adapts across different parts of the human skeleton remains a challenge. Moreover, the incorporation of these learnable activations increases the computational cost per layer. The use of spline-based activations further contributes to memory overhead, as storing and evaluating function approximations demand additional resources. For large-scale datasets, this can lead to substantial memory consumption, particularly when training deeper networks. Therefore, further improvements are necessary to enhance computational efficiency, interpretability, and robustness, ensuring broader applicability of PoseKAN.

\end{document}